\documentclass[sigconf]{acmart}

\AtBeginDocument{%
  \providecommand\BibTeX{{%
    \normalfont B\kern-0.5em{\scshape i\kern-0.25em b}\kern-0.8em\TeX}}}

\newcounter{BalanceAtReference}
\newcounter{ReferenceIndexForBalancing}

\makeatletter

\global\@ACM@balancefalse

\global\@ACM@balancefalse

\def\@balancelastpageonce{%
  \ifnum\value{ReferenceIndexForBalancing}=\value{BalanceAtReference}
    \newpage
  \else
    \relax
  \fi
  \stepcounter{ReferenceIndexForBalancing}
}
\pretocmd{\bibitem}{\@balancelastpageonce}
  {} 
  {\@latex@error{Patching \bibitem failed}{\@ehd}}

\makeatother

\usepackage{amsmath}
\usepackage{amsfonts}
\usepackage[linesnumbered,ruled,vlined]{algorithm2e}
\usepackage{graphicx} 
\usepackage{booktabs}
\usepackage{array}
\usepackage{enumitem}
\usepackage{amsthm}
\usepackage{algpseudocode}
\usepackage[switch]{lineno}
\usepackage[utf8]{inputenc}
\usepackage{eqparbox}
\usepackage[nopar]{lipsum}
\usepackage{multirow}
\usepackage{makecell}
\usepackage{xcolor}
\usepackage{hhline} 
\usepackage{microtype}
\usepackage{diagbox}
\usepackage{longtable} 
\usepackage{adjustbox}
\usepackage{mdframed}
\usepackage{framed,color} 
\usepackage{balance}

\newtheorem*{proof*}{Proof}
\newtheorem*{claim*}{}

\newcolumntype{C}[1]{>{\centering\let\newline\\\arraybackslash\hspace{0pt}}m{#1}}

\begin{document}
\fancyhead{}

\title{Lighter And Better: Towards Flexible Context Adaptation For Retrieval Augmented Generation}

\author{
    Chenyuan Wu
}
\authornote{These authors contribute equally to this paper.}
\affiliation{
\institution{University of Science and Technology of China \country{}}} 
\email{wu.chenyuan@ustc.edu.cn} 

\author{
    Ninglu Shao
}
\authornotemark[1]
\affiliation{
\institution{Gaoling School of AI, Renmin University of China \country{}}} 
\email{shao.ninglu@ruc.edu.cn}

\author{
    Zheng Liu
}
\authornotemark[1]
\affiliation{
\institution{Beijing Academy of Artificial Intelligence \country{}}} 
\email{zhengliu1026@gmail.com} 

\author{
    Shitao Xiao
}
\affiliation{
\institution{Beijing Academy of Artificial Intelligence \country{}}} 
\email{stxiao@baai.ac.cn} 


\author{
    Chaozhuo Li
}
\affiliation{
\institution{Beijing University of Posts and
Telecommunications \country{}}} 
\email{cli@bupt.edu.cn} 

\author{
    Defu Lian
}
\affiliation{
\institution{University of Science and Technology of China \country{}}} 
\email{lian.defu@ustc.edu.cn}

\renewcommand{\shortauthors}{Wu, Shao, and Liu, et al.}



\begin{abstract}
    The existing Retrieval-Augmented Generation (RAG) systems face significant challenges in terms of cost and effectiveness. On one hand, they need to encode the lengthy retrieved contexts before responding to the input tasks, which imposes substantial computational overhead. On the other hand, directly using generic Large Language Models (LLMs) often leads to sub-optimal answers, while task-specific fine-tuning may compromise the LLMs' general capabilities. 
    To address these challenges, we introduce a novel approach called \textbf{FlexRAG} (\underline{Flex}ible Context Adaptation for \underline{RAG}). In this approach, the 
    retrieved contexts are compressed into compact embeddings before being encoded by the LLMs. Simultaneously, these compressed embeddings are optimized to enhance downstream RAG performance.   
    A key feature of FlexRAG is its flexibility, which enables effective support for diverse compression ratios and selective preservation of important contexts.  
    Thanks to these technical designs, FlexRAG achieves superior generation quality while significantly reducing running costs. Comprehensive experiments on various question-answering datasets validate our approach as a cost-effective and flexible solution for RAG systems. 
\end{abstract}

\keywords{Retrieval Augmented Generation, Large Language Models, Question Answering, Context Compression and Optimization}

\maketitle 

\vspace{-5pt}
\section{Introduction}
Large language models (LLMs) are growing as a general foundation of artificial intelligence. However, the existing LLMs are still limited by incomplete and outdated knowledge due to their static nature, and this limitation is particularly pronounced when dealing with knowledge-intensive tasks \cite{sun2024trustllm,liu2023trustworthy,ji2023survey}. To mitigate this limitation, people resort to retrieval-augmented generation (RAG). With proper information retrieved from external databases, the generation process can be conducted on top of knowledge-grounded contexts. Consequently, it substantially contributes to LLMs' generation quality in terms of truthfulness and credibility \cite{gao2023retrieval}. In recent years, influential prototyping systems, e.g., WebGPT, SearchGPT \cite{searchgpt}, Perplexity \cite{perplexity}, and developing frameworks, such as Llama-Index, Lainchain, are continuously proposed by the community, facilitating both application and research in this area. 


\subsection{The Challenges}
Despite the widespread popularity, existing RAG systems still face significant challenges, particularly in terms of running costs and effectiveness. Firstly, RAG systems often require processing lengthy contexts to handle knowledge-intensive tasks. For instance, solving multi-hop QA tasks may involve working with a series of correlated documents \cite{trivedi2022musique,ho2020constructing,yang2018hotpotqa}, while general language modeling tasks may call for iterative retrieval of diverse knowledge sources \cite{jiang2023active,asai2023self}. In such situations, it will take substantial computation costs in order to encode the lengthy contexts for LLMs. Secondly, the RAG systems can be limited by their answer quality if generic LLMs are directly utilized. This limitation is especially evident for many public models of moderate scale, which often struggle to effectively utilize the retrieved knowledge, particularly in complex and noisy contexts \cite{gao2023retrieval}. While task-specific fine-tuning can improve the answer quality \cite{zhang2024raft}, it may come at the cost of reduced instruction-following capabilities and diminished performance on other general tasks \cite{luo2023empirical}.

\begin{figure*}[t]
\centering
\includegraphics[width=0.99\textwidth]{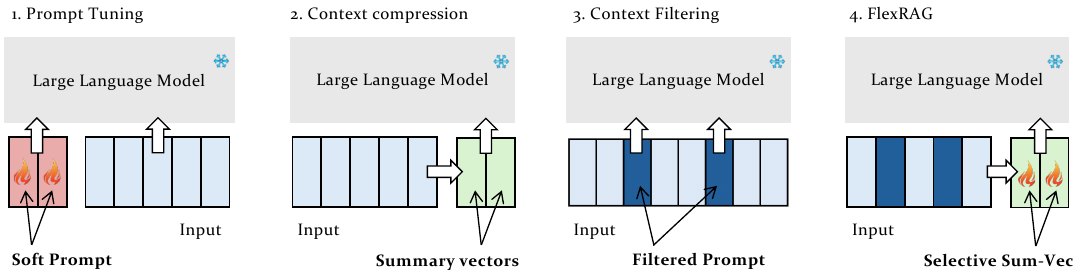}
\caption{Comparison of related techniques. 1) \textit{Context compression}: token embeddings are compressed into compact summary vectors. 2) \textit{Context filtering}: important token embeddings are filtered from the input prompt. 3) \textit{Prompt tuning}: soft-prompt is learned to improve the downstream task. 4) \textit{FlexRAG}: unifying all functions in one framework, with compressive embeddings (summary vectors) down-sampled by importance (filtering) and learned to optimize the RAG performance (prompt tuning).} 
\label{fig:1} 
\end{figure*} 

\vspace{-5pt}
\subsection{Our Approach} 
To address the above challenges, we propose a novel approach in this paper, called \textbf{FlexRAG} (shown in Figure \ref{fig:1}. 4). It transforms the retrieved contexts into compact and more usable forms, which substantially improves the cost-effectiveness of RAG systems. 

First of all, FlexRAG helps to \textbf{reduce the running cost of RAG}. Essentially, FlexRAG pre-encodes external documents into \textit{compressive embeddings} during the offline stage and performs \textit{down-sampling} of the compressive embeddings when corresponding documents are retrieved for specific RAG tasks. Since the down-sampled token embeddings are significantly shorter compared to directly tokenized documents, this approach substantially reduces the computation cost for RAG systems. A key characteristic of FlexRAG is its \textbf{flexibility}. Previous methods typically perform static compression of input context based on predefined compression ratios  \cite{chevalier2023adapting,mu2024learning,ge2023context}. In contrast, our approach supports \textbf{arbitrary compression ratios} specified by the user, and enables \textbf{selective compression of the contexts} based on their importance in specific scenarios. This means that the critical parts of the context are preserved as much as possible, while the less important parts are assigned with large compression ratios. Consequently, useful information within the context can be better presented for downstream RAG tasks. 

Secondly, FlexRAG \textbf{optimizes the performance of RAG} in a compatible way. In our work, a two-stage training workflow is designed for FlexRAG. In the first stage, we employ task-generic pre-training using a generic corpus, like RedPajama \cite{together2023redpajama}, which establishes the preliminary alignment between the compression module and downstream LLM. In the second stage, we perform task-specific fine-tuning using various instruction-tuning datasets, which optimizes the answer quality for downstream RAG tasks. Throughout the entire training process, the compression module remains learnable while the LLM parameters are kept fixed. As no modification is made to the LLM's original parameters, we can optimize the performance in RAG without compromising the performance in other general tasks. 


We perform comprehensive empirical studies using a variety of question-answering datasets. In our experiments, FlexRAG exhibits three key advantages. 1) \textbf{Superior cost-effectiveness}, where substantial improvements can be achieved over generic LLMs and other context compression methods with significantly reduced costs. 2) \textbf{Flexibility of usage}, as it effectively supports various compression ratios and compression methods. 3) \textbf{General usability}, as the competitive performance can be well-preserved across various datasets and working conditions. These results validate FlexRAG as an effective and economical component for RAG systems. 

To summarize, the following technical contributions are highlighted for our paper:
\begin{itemize}
    \item We propose a novel method, FlexRAG, for compressive and optimized adaptation of the retrieved contexts for RAG.  
    \item FlexRAG realizes flexible compression of the retrieved contexts by various ratios, and enables selective preservation of useful information leveraging estimated importance. 
    \item We design a two-stage training workflow for FlexRAG. By making sufficient utilization of available data, it effectively enhances the downstream performance of RAG. 
    \item We perform comprehensive experiments with a variety QA datasets, whose result verifies the cost-effectiveness, flexibility, and general usability of FlexRAG. 
\end{itemize}

\section{Related Works}
In this section, the related works are discussed from three perspectives: 1) retrieval-augmented generation, 2) context compression, 3) fine-tuning for RAG optimization. 

\subsection{Retrieval-augmented Generation}
Retrieval-Augmented Generation has emerged as a crucial paradigm for language models~\cite{lewis2020retrieval}, particularly with the rise of LLMs. A typical RAG system consists of two components: retrieval tools that access external databases and a language model that generates knowledge-grounded content based on the retrieval results. By introducing relevant knowledge, RAG significantly enhances the truthfulness and credibility of LLM-generated outputs, making it a valuable approach for mitigating hallucinations \cite{ji2023survey,liu2023trustworthy}. Additionally, RAG offloads internal knowledge to external memory, contributing to improved cost-effectiveness of LLMs \cite{borgeaud2022improving,izacard2023atlas}. 

In recent years, RAG has become a significant research focus in both academia and industry. Researchers have continuously proposed advanced architectures beyond the basic direct prompting approach, such as fusion-in-decoder \cite{izacard2020leveraging} and internal knowledge injection \cite{borgeaud2022improving}, which facilitate the effective use of the retrieved knowledge. Meanwhile, the training of retriever and generator has been improved from simple independent training to more advanced forms of joint training \cite{izacard2023atlas,shi2023replug,zhang2023retrieve}, offering users the flexibility to select the most suitable method for their specific applications. Moreover, designing appropriate mechanisms for RAG, such as determining when and where to apply retrieval augmentation and which information to retrieve, is crucial. Significant progress has been made in this field, with approaches like uncertainty-based methods (e.g., FLARE \cite{jiang2023active}), self-prompting methods (e.g., ToolFormer \cite{schick2024toolformer}, Self-RAG \cite{asai2023self}), and reflection-based methods (e.g., ReAct \cite{yao2022react}, Reflexion \cite{shinn2024reflexion}) proposed for more effective control of RAG in complex real-world scenarios. Additionally, RAG's application has been actively explored beyond traditional question-answering \cite{lewis2020retrieval}, such as long-context modeling \cite{xu2023retrieval}, in-context learning \cite{zhang2023retrieve,wang2023learning}, code generation \cite{wang2024coderag}, and multi-modal processing \cite{yasunaga2022retrieval}, etc.



\subsection{Context Compression}

Running costs pose a significant challenge for RAG systems, primarily due to the need to encode lengthy retrieved contexts. To address this, an important strategy involves compressing the retrieved contexts before they are processed by downstream LLMs. In line with this approach, various compression techniques have been developed in recent years. 
One notable work is Gist~\cite{mu2024learning}, which implicitly compresses long contexts using a small number of Gist tokens. Similarly, ICAE~\cite{ge2023context} fine-tunes LLMs as specialized context compressors through LoRA, while AutoCompressor~\cite{chevalier2023adapting} integrates compression learning with the autoregressive language modeling process. 
In addition to these implicit methods, another line of research focuses on explicit filtering of contexts, where less important tokens are removed directly.  A representative study in this area is LLMLingua \cite{jiang2023llmlingua,jiang2023longllmlingua}, which uses coarse-to-fine approaches to compress contexts based on given budgets. Besides, RECOMP \cite{xu2023recomp} presents both extractive and abstractive compressor, which selects useful sentences and generates summaries from long-contexts, respectively. While explicit methods are generally agnostic to downstream LLMs and therefore more practical, they may suffer from higher compression loss due to the over-removal of input tokens.

\subsection{RAG Fine-tuning}
While RAG systems can be constructed using off-the-shelf retrievers and LLMs, this native approach often results in sub-optimal performance. Issues may include insufficient utilization of retrieved knowledge, vulnerability to retrieval noise, and misalignment with the required answer format or human preferences \cite{gao2023retrieval}. Consequently, continual fine-tuning is often necessary to enhance RAG performance. In the simplest scenarios, retrievers and LLMs can be independently fine-tuned using their respective training data \cite{zhang2024raft}. However, more advanced approaches involve joint training of retrievers and LLMs. In these cases, retrievers are optimized to select contexts that are more conducive to the LLM's processing, while LLMs are trained to adapt to the specific contexts provided by the retrievers \cite{izacard2023atlas,shi2023replug}. 
Nevertheless, fine-tuning RAG systems isn't without its drawbacks. It can lead to a reduction in the general capacity of LLMs, a phenomenon known as catastrophic forgetting. To mitigate this issue, parameter-efficient fine-tuning (PEFT) is often employed, where only specialized learnable adapters are fine-tuned, leaving the LLM's original parameters intact \cite{ding2023parameter}. Among various PEFT methods, prompt-tuning is particularly effective in minimizing the impact on LLMs \cite{lester2021power,liu2021p}. In this approach, adaptation modules, i.e. soft prompts, are introduced as external components. Following this spirit, our encoder is designed, which produces both compressive and RAG-optimized embeddings for downstream LLMs. 


\begin{figure*}[t]
\centering
\includegraphics[width=0.95\textwidth]{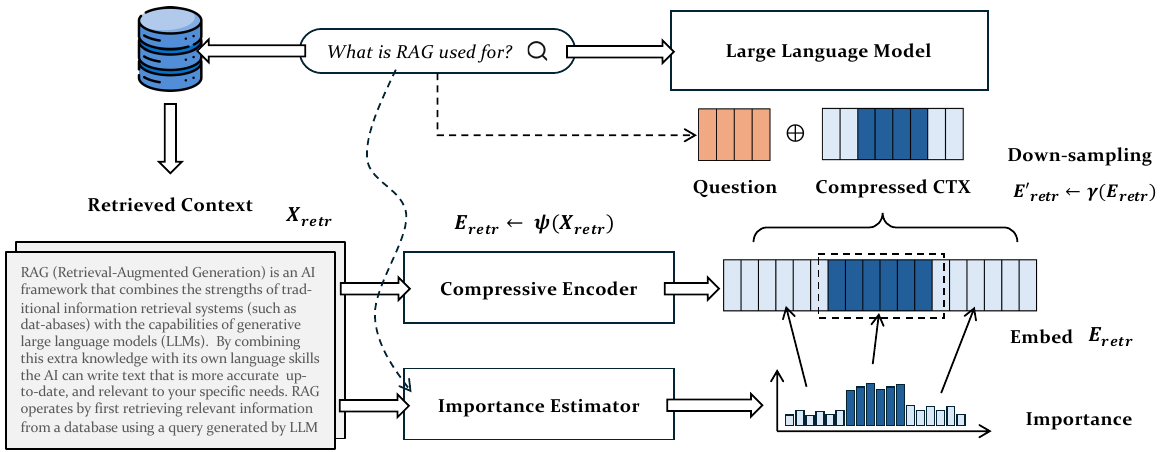}
\vspace{-10pt}
\caption{Architecture of FlexRAG. It transforms the retrieved contexts $\mathbf{X}_{retr}$ into compressive embeddings $\mathbf{E}_{retr}$ using the compressive encoder. With estimated importance, it down-samples $\mathbf{E}_{retr}$ into $\mathbf{E}'_{retr}$ as the compressed context for RAG.} 
\vspace{-5pt} 
\label{fig:2} 
\end{figure*}

\section{Methodology}
In this section, we delve into the technical aspects of FlexRAG. We begin by formulating the problem of context adaptation for RAG. Following this, we introduce the architecture of FlexRAG, focusing on its two basic components: the compressive context adapter and the selective compression mechanism. Finally, we introduce the optimization of FlexRAG using various types of data.  

\subsection{Problem Formulation}
Retrieval-Augmented Generation (RAG) is a specialized working paradigm for Large Language Models (LLMs), which is designed to facilitate knowledge-intensive tasks, such as question-answering and knowledge-grounded dialogue systems. It presents the task's prompt (denoted as $\mathrm{X}_{task}$) together with the retrieved context (denoted as $\mathrm{X}_{retr}$) as the input for LLM, based on which the ground-truth answer (denoted as $\mathrm{X}_{ans}$) is predicted. The retrieved context is expected to include necessary knowledge to the task, such that the ground-truth answer can be better predicted. In spite of widespread popularity, the directly application of LLMs for RAG is constrained by suboptimal performance and high computational costs. To address these challenges, we propose \textbf{Flexible Context Adaptation for RAG} with the following objectives: 1) the retrieved context is compressed into a concise form: $|\phi(\mathrm{X}_{retr})| < |\mathrm{X}_{retr}|$ ($\phi(\cdot)$ stands for the compressor); 2) the compressed context helps to deliver optimized performance for RAG. These objectives can be generalized and formulated as the following optimization problem:  
\begin{equation}
\begin{split}
    & \max\limits_{\phi} ~ 
    P_{LLM}(\mathrm{X}_{ans}|\phi(\mathrm{X}_{retr}),\mathrm{X}_{task}) \\
    & ~~ s.t. ~ |\phi(\mathrm{X}_{retr})| = k, ~~\text{where } k < |X_{retr}|
\end{split}
\end{equation} 
In other words, we aim to realize the \textbf{optimal compression}, where the compressed context of the predefined size $k$ can maximize the generation likelihood of the ground-truth answer. Additionally, the compressor is expected to achieve \textbf{flexible compression}, allowing the retrieved context to be compressed to any length within the original context length $|X_{retr}|$.


\subsection{Architecture}
The workflow of FlexRAG consists of the following steps (Figure \ref{fig:2}). First, the retrieved context is tokenized and jointly encoded as a sequence of embeddings, denoted as $\mathbf{E}_{retr}$, using a specialized context encoder $\psi(\cdot)$: $\mathbf{E}_{retr} \leftarrow \psi(X_{retr})$. Next, the well-encoded embeddings $\mathbf{E}_{retr}$ are down-scaled by a sampling function $\gamma(\cdot)$: 
\begin{equation}
    \mathbf{E}'_{retr} \leftarrow \gamma(\mathbf{E}_{retr}, k), ~~~~
    \text{where}~~~~ |\mathbf{E}'_{retr}| = k,
\end{equation}
In this place, $k$ indicates the predefined size of compressed context. The down-scaled embeddings $\mathbf{E}'_{retr}$ serve as compact yet informative representations of the retrieved context, which are passed to the downstream LLM for retrieval-augmented generation. 

The above workflow consists of two basic modules. The first is the \textbf{compressive encoder}, which implements the encoding function $\psi(\cdot)$ to transform the retrieved context into informative and flexible-to-sample embeddings. The second is the \textbf{importance estimator}, which assesses the importance of each part of the context. Based on the estimation results, selective compression is conducted through down-sampling, i.e. $\gamma(\cdot)$, allowing for the preservation of the most critical information with higher emphasis.  




\subsection{Compressive Encoder}
As described, the compressive encoder transforms the tokenized retrieved contexts $\mathbf{X}_{retr}$ into compressive embeddings $\mathbf{E}_{retr}$, which can be flexibly down-sampled to provide a high-quality compression of the original input. As a result, the realization of compressive encoder is guided by the following considerations. 

First, to serve as compressive representations of input, each element of $\mathbf{E}_{retr}$ needs to fully capture the information for its nearby context. This requires a highly \textbf{expressive} encoding backbone to generate rich-semantic embeddings. To achieve this goal, we exploy LLMs as the foundation of our compressive encoder.

Second, it needs to ensure a \textbf{seamless connection} between the compressive embeddings and the downstream LLM. To facilitate this objective, the compressive embeddings must resemble the input token embeddings of the downstream LLM. With this consideration, we employ the same backbone as the downstream LLM. Besides, we choose to leverage the first-$n$ layers instead of the entire LLM, considering that the intermediate embeddings from the mid-layers are more similar to the LLM's token embeddings. 

Third, the length of retrieved context is likely to exceed the window size of compressive encoder, making it inevitable to chunk the input and encode each segment individually. However, it has to avoid over-chunking so as to maintain the coherence of input. Therefore, the chunking size is expanded to the maximum extent in each specific scenario.  

Therefore, the compressive encoding can be formulated as the following workflow: 
\begin{equation}
    \psi(\mathbf{X}_{retr}) 
    \rightarrow
    \bigl[
    \mathrm{LLM}^{e}_{:\mathrm{n}}(\mathbf{X}_{retr}^1), ~ ...  ~ ,
    \mathrm{LLM}^{e}_{:\mathrm{n}}(\mathbf{X}_{retr}^m) 
    \bigr]
    \rightarrow 
    \mathbf{E}_{retr} 
\end{equation}
In this place, $\mathrm{LLM}^{e}$ is the LLM backbone employed for compressive encoding, while ``$:\mathrm{n}$'' indicates that the first-n layers are utilized. 



\subsection{Selective Compression Mechanism}
Once the retrieved contexts are encoded as compressive embeddings, selective compression is applied, which produces the compressed context for RAG through down-sampling. It emphasizes the useful information to RAG tasks, where the related contexts are assigned with a high sampling ratio. In contrast, it neglects the less useful information, whose related contexts are assigned with a small sampling ratio. With this processing, the useful information can be better preserved from compression. 

Selective compression calls for accurate estimation of context importance. In our work, we propose two alternative approaches to achieve this goal, providing users with the flexibility to choose the most suitable option for their specific applications.


\subsubsection{Token-level estimation}
The first alternative estimates context importance on the token basis, which is a popular principle adopted by many studies \cite{jiang2023llmlingua,pan2024llmlingua}. Given the input prompt of RAG task $X_{task}$, the importance tokens within the retrieved contexts $X_{retr}$ are generally favored by the LLM, leading to relatively higher generation likelihood compared to other less useful tokens. Based on this principle, we introduce the following relationship as an approximate indicator of token importance:  
\begin{equation}\label{eq:likelihood}
    \text{for} ~~ x_i \in X_{retr}: ~~
    w_i \leftarrow P_{LLM}(x_i|{X}_{task}, X_{retr}[:x_i] ) 
\end{equation}
where $X_{retr}[:x_i]$ represents the prefix of $x_i$. In other words, $w_i > w_j$ if $x_i$ is more important than $x_j$.  
Despite simplicity, the above indicator can basically identify useful contexts as demonstrated by related works \cite{jiang2023llmlingua,pan2024llmlingua}. However, it might suffer from incoherence and broken semantic as discrete tokens are sampled from context.   

\subsubsection{Sentence-level estimation}
The second alternative estimates the importance for each sentence within the retrieved contexts. To this end, we employ an ad-hoc model to estimate the sentence's relevance to the task's prompt, where the relevance score is used as the importance. Although the original retriever which produce the retrieved contexts is a desirable option, it is not always always in practice. Therefore, we leverage a general purpose embedder, such as E5 \cite{wang2022text} and BGE \cite{bge_embedding}, as the relevance oracle, denoted as $\mathcal{M}$. Therefore, the importance is computed as:
\begin{equation}\label{eq:embedding}
    \text{for}~~ sent_i \in X_{retr}: ~~
    w_i \leftarrow \mathcal{M}(X_{task}, sent_i),
\end{equation}
where $sent_i$ stands for the $i$-th sentence in the retrieved contexts. Compared to the token-level method, the sentence-level approach is able to better maintain  semantic coherence, as the retrieved contexts are down-sampled on a sentence basis.


\subsubsection{Compression ratio allocation} 
Although the estimated importance is positively correlated with the usefulness of context, it serves more as an indicator of relative relationships rather than a direct basis for sampling ratios. To address this, we propose a stepped scheme for allocating the sampling ratios.
This method partitions the retrieved contexts into groups, where higher-priority groups receive a greater sampling ratio. The allocation process is straightforward which involves three simple steps. 
First, we rank different parts of the contexts based on their estimated importance (by tokens with the token-level estimation, or by sentences with the sentence-level estimation). Next, we introduce $k$ groups: $g_1$, $g_2$, ... , $g_k$, with increasing priorities. We also define sampling ratio for these $k$ groups: $w_1$, $w_2$, ... , $w_k$, in an ascending order. Finally, we make linear allocation of the contexts to the $k$ groups to ensure that the following relationship holds: 
\begin{equation}
    w_1*n_1 + w_2*n_2 + ... w_k*n_k = \alpha*n
\end{equation}
Here, $n_i$ and $n$ represent the length for the $g_i$ and $X_{retr}$, respectively, while $\alpha$ denotes the required compression ratio. In the simplest case where a binary partition is made (i.e., forming a low-priority group $g_1$ and a high-priority group $g_2$ are formed), the context allocation can be directly calculated once the sampling ratios are determined. 



\subsection{Training Workflow}
The training process takes place to optimize the compressive encoder, enabling it to generate high-quality compressed contexts that enhance RAG performance. Given the abundance of unlabeled data (e.g., general corpora like Pile, RedPajama) and the limited availability of labeled data (e.g., for question answering), we design a \textbf{two-stage training workflow} to fully optimize the model based on the accessible data resource. First, we perform auto-regressive \textbf{pre-training} on the unlabeled data, where language modeling is conducted based on the compressed contexts. In this stage, the following training objective is maximized: 
\begin{equation}
    \max \sum\nolimits_{x_i \in X_{pre}} P_{LLM} (x_i|\phi(X_{pre}[:x_i])). 
\end{equation}
Here, $X_{pre}$ stands for a sample of unlabeled data for pre-training, $x_i$ is the $i$-th token, and $\phi(X_{pre}[:x_i])$ is the compression of $x_i$'s prefix. With the first stage of training, the connection is established between the compressive encoder and the downstream LLM. Next, we move on to perform fine-tuning based on label QA datasets. For each training sample, the ground-truth answer $X_{ans}$ is predicted based on the question $X_{q}$ and the compressed retrieved contexts $\phi(X_{retr})$ (i.e. relevant docs to the question). As a result, we can formulate the following objective function: 
\begin{equation}
    \max \sum\nolimits_{x_i \in X_{ans}} P_{LLM} (x_i|\phi(X_{retr}), X_q, X_{ans}[:x_i]).   
\end{equation}
Thanks to the second stage of training, the compressive encoder can be further enhanced to optimize the RAG performance. 

The training process is further \textbf{enhanced in two ways}. First, we introduce {two-stream processing} during pre-training. This involves encoding all chunks of each input at the very beginning (encoding stream) and then performing auto-regressive decoding based on the pre-encoded contexts (decoding stream). This approach allows for parallelized auto-regressive language modeling, making it more sample-efficient than traditional recurrent method \cite{chevalier2023adapting}. Second, we {randomly sample} the compressive embeddings using a {dynamic rate} during training. In other words, selective compression is only made during inference. This approach lets the entire output of the encoder to be trained as compressive embeddings; meanwhile, it also enables the model to flexibly accommodate various compression ratios.   



\section{Experiments}
Our experiments are dedicated to the following research questions. \textbf{RQ. 1} Can FlexRAG bring forth cost-effective compression results for general RAG tasks. \textbf{RQ. 2} Can FlexRAG flexibly support diverse compression ratios and compression methods. \textbf{RQ. 3} Extended analysis of various aspects, including FlexRAG's robustness to different working conditions and the impact of each technical factor. 


\subsection{Settings}
\subsubsection{Datasets} The experiments focus on evaluating RAG performance, where two types of tasks are used: Long-sequence Multi-doc QA (LMQA) and conventional Open-Domain QA (ODQA). For LMQA, the retrieved contexts consist of multiple long documents, whose entire length is usually longer than the LLM's window size. We include the following datasets for LMQA: HoptpotQA, 2WikiMQA, Musique, where we use the curated version offered by LongBench \cite{bai2023longbench}. The retrieved contexts have been well-presented in these datasets, thus no additional retriever is needed. For ODQA, the retrieved contexts consist of short passages from Wikipedia corpus, whose entire length is usually within the context window of LLM. We include the following datasets for ODQA: Natural Questions (NQ), PopQA, and TriviaQA, where we use the curated version offered by KILT~\cite{petroni2020kilt}. The retrieved contexts are not presented in these datasets, therefore, we employ various retrievers to undertake this role in our experiment. Following the requirements from LongBench and KILT, we use F1 and Exact Match (EM) as the metrics for LMQA and ODQA, respectively. 

\subsubsection{Baselines} We make comparison with various types of popular baselines in our experiment. First, we introduce two methods which employ LLaMA-2-7B (chat) for question answering: 1) Llama (retrieval), which directly makes use of the retrieved contexts without compression, and Llama (w/o retrieval) which answers the question without using retrieved contexts. Second, we include two types of context compression methods. One is the context compression methods, which contains ICAE~\cite{ge2023context} and AutoCompressor \cite{chevalier2023adapting}. Both methods generate summary vectors as the compressed inputs for RAG. The other one is the context filtering methods, including LLMLingua \cite{jiang2023llmlingua},  LongLLMLingua \cite{jiang2023longllmlingua}, and TF-IDF (as implemented by Gist~\cite{mu2024learning}). These methods filtering important tokens from the retrieved contexts for RAG tasks. 


\begin{table*}[h]
    \centering
    \resizebox{1.0\linewidth}{!}{
    \begin{tabular}{l|c|cccc|cccc}
    \toprule
        & & \multicolumn{4}{c|}{LMQA} & \multicolumn{4}{c}{ODQA} \\ 
        \cmidrule(lr){1-2}\cmidrule(lr){3-10}
        \cmidrule(lr){3-10}
        Method & CP. Ratio & HotpotQA & 2WikiMQA & Musique & Average & NQ & PopQA & TriviaQA & Average \\
    \midrule
        Llama (w/o retrieval) & -- & 19.07	& 27.78 & 5.65 & 17.50 & 14.17	& 18.01	& 53.67	& 28.62 \\
        Llama (w. retrieval) & -- & 27.20 & 32.21 & 7.61 & 22.34 & 25.13 & 31.04 & 55.59 & 37.25 \\
    \midrule
        ICAE \cite{ge2023context} & 8 $\times$ & 19.56 & 25.07 & 5.73	& 16.79	& 10.53	& 4.44 & 8.21 & 7.73\\
        AutoCompressor \cite{chevalier2023adapting} ~~ & 8 $\times$ & 13.80 & 17.31 & 7.05 & 12.72 & 13.20 & 17.78 & 49.55 & 26.84\\
        TF-IDF \cite{mu2024learning} & 8 $\times$ & 19.20 & 24.77 & 6.28	& 16.75	& 11.49	& 14.05	& 46.73	& 24.09\\
        LLMLingua \cite{jiang2023llmlingua} & 8 $\times$ & 21.07 & 26.40 & 5.46 & 17.64 & 10.22 & 11.92 & 35.75 & 19.30\\
        LongLLMLingua \cite{jiang2023longllmlingua} ~~ & 8 $\times$ & 21.55 & 24.77 & 7.15	& 17.82	& 18.15	& 22.74	& 52.77	& 31.22\\
        LLMLingua-2 \cite{pan2024llmlingua} & 8 $\times$ & 29.51 & 26.37 & 11.89 & 22.59 & 14.03 & 16.67 & 44.19 & 24.96\\
                
    \midrule
        FlexRAG w/o SC. & 8 $\times$ & 33.81 & 38.76 & 12.56 & 28.38 & 31.44 & 24.55 & 65.07 & 40.35 \\
        FlexRAG w. SC. & 8 $\times$ & \textbf{36.30} & \textbf{39.15} & \textbf{14.33} & \textbf{29.93} & \textbf{33.45} & \textbf{35.96} & \textbf{66.71} & \textbf{45.37} \\
    \bottomrule
    \end{tabular}
    }
    \caption{Cost-effectiveness Analysis on LMQA and ODQA (FlexRAG w. SC. stands for FlexRAG with selective compression).} 
    \vspace{-15pt}
    \label{tab:main}
\end{table*}

\begin{table}[h]
    \centering
    \resizebox{1.0\columnwidth}{!}{
    \begin{tabular}{ccccc}
    \toprule
    Model & CP. Ratio & EM & CUDA Time (s) & TFLOPs\\
    \midrule
    Llama (w.r.)  & 1 $\times$ & 37.25 & 7.78 & 14.17 \\
    \midrule
    \multirow{4}{*}{FlexRAG} & 2 $\times$ & 47.23 & 4.97 & 10.48\\
     & 4 $\times$ & 47.25 & 3.13 & 7.03 \\
     & 8 $\times$ & 45.37 & 2.48 & 5.39\\
     & 16 $\times$ & 38.93 & 2.20 & 4.59\\
    
    \bottomrule
    \end{tabular}
    }
    \caption{Efficiency analysis using CUDA Time and TFLOPs. Compression ratios are varied from 2$\times$ to 16$\times$. 
    Experiments are performed on ODQA with EM as the quality metric.} 
    \vspace{-10pt}
    \label{tab:efficiency}
\end{table}

\subsubsection{Implementations} We initialize FlexRAG with the first 8 layers of LLaMA-2-7B (chat), and 
we leverage LLaMA-2-7B (chat) \cite{touvron2023llama-b} as our downstream LLM. This ensures a fair comparison with the baselines and maintains the economical running of the experiments. The pre-traininig is performed with 90K sampled instances from Redpajama \cite{together2023redpajama} and 10K training instances from LongAlpaca \cite{chen2023longlora}, while the fine-tuning is performed with 10K sampled instances from a blend of HotpotQA~\cite{yang2018hotpotqa} and Natural Questions dataset \cite{kwiatkowski2019natural}. During training, the compression ratios are randomly sampled from 1, 2, 4, and 8. The training takes place on a Nvidia 8×A800 GPU machine. By default, we use BGE-EN-large \cite{bge_embedding} as the retriever, where the top-5 documents are returned during the testing stage. Meanwhile, FlexRAG's compression is made by sentence-level importance estimation and selection. 


\subsection{Cost-Effectiveness Analysis}
\subsubsection{Primary results}
We first evaluate the primary question answering performance under the default setting, where a uniform compression ratio (CP. Ratio) of 8$\times$ is applied. For LMQA, all retrieved contexts are confined within 32K tokens, allowing them to be fully utilized by the downstream LLM after compression. In contrast, Llama (w. retrieval) must truncate the retrieved contexts to fit within the 4K window of Llama-2, as implemented by Longbench \cite{bai2023longbench}. We compare two variants of our method: FlexRAG w/o SC, which disables selective compression and uniformly down-samples the retrieved contexts at an interval of 8 tokens, and FlexRAG w. SC, the default method using selective compression. The following observations can be drawn from the experiment results in Table \ref{tab:main}.

First, FlexRAG demonstrates superior performance in the experiment. Even without using selective compression, FlexRAG w/o SC already outperforms all baselines with notable advantages. With the enhancement from selective compression, FlexRAG's performance is further improved, which leads to the optimal question answering quality across all datasets. The above result preliminarily validates the effectiveness FlexRAG, indicating that the context compression and RAG optimization are realized simultaneously.  

Second, FlexRAG's effectiveness is more pronounced on ODQA tasks, whose retrieved contexts are much shorter than those from LMQA, e.g., it notably improves upon the best of the compression baselines (LongLLMLingual) from 18.15 to 33.45 on PopQA. This is because when retrieved contexts are concise, they are more likely to suffer from information loss once compressed. It is a more challenging task to handle the compression on ODQA, and the increased challenge expands the gap between FlexRAG and baselines. 

Third, Llama (w. retrieval) falls behind many compression baselines in LMQA, whereas it outperforms all of them in ODQA. As mentioned, the compression baselines can fully leverage the retrieved contexts with a uniform compression ratio 8$\times$; while Llama (w. retrieval) has to truncate the retrieved contexts, which incurs information. However, the retrieved context is concise on ODQA, where no truncation is needed for  Llama (w. retrieval); by comparison, the compression baselines will not leverage any extra information, but suffer from the information loss caused by compression.

\begin{table*}[h]
    \centering
    \resizebox{1.0\linewidth}{!}{
    \begin{tabular}{l|c|cccc|cccc}
    \toprule
        &  & \multicolumn{4}{c|}{LMQA} & \multicolumn{4}{c}{ODQA} \\
        \cmidrule(lr){1-2}\cmidrule(lr){3-10}
        Method & CP. Ratio & HotpotQA & 2WikiMQA & Musique & Average & NQ & PopQA & TriviaQA & Average \\
    \midrule
        Llama (w/o retrieval) & -- & 19.07	& 27.78 & 5.65 & 17.50	& 14.17	& 18.01	& 53.67	& 28.62 \\
        Llama (w. retrieval) & -- & 27.20 & 32.21 & 7.61 & 22.34 & 25.13 & 31.04 & 55.59 & 37.25 \\
    \midrule
        \multirow{4}{*}{FlexRAG w/o SC.} & 1 $\times$ & 30.83 & 35.01 & 12.86 & 26.23 & 37.01 & 33.92 & 66.93 & 45.95\\
        & 2 $\times$ & 32.06 & 35.72 & 12.02 & 26.60 & 36.62 & 31.03 & 67.74 & 45.13 \\
        & 4 $\times$ & 31.23 & 34.19 & 11.83 & 25.75 & 35.28 & 28.09 & 67.10 & 43.49 \\
        & 8 $\times$ & 33.81 & 38.76 & 12.56 & 28.38 & 31.44 & 24.55 & 65.07 & 40.35 \\
    \midrule
        \multirow{3}{*}{FlexRAG w. SC.} & 2 $\times$ & 34.31 & 37.41 & 13.28 & 28.33 & \textbf{37.19} & 36.10 & 68.41 & 47.23 \\
         & 4 $\times$ & 35.06 & 36.40 & 14.93 & 28.80 & 36.02 & \textbf{37.15} & \textbf{68.59} & \textbf{47.25} \\
         & 8 $\times$ & \textbf{36.30} & \textbf{39.15} & \textbf{14.33} & \textbf{29.93} & 33.45 & 35.96 & 66.71 & 45.37 \\
    \bottomrule
    \end{tabular}
    }
    \caption{Flexibility analysis on LMQA and ODQA using different compression ratios (1$\times$, 2$\times$, 4$\times$, 8$\times$).}
    \vspace{-15pt}
    \label{tab:comp_ratio}
\end{table*}

\begin{table*}[h]
    \centering
    \resizebox{1.0\linewidth}{!}{
    \begin{tabular}{c|c|c|ccc|cccc}
    \toprule
    \multirow{2}{*}{Method} & \multirow{2}{*}{Estimator} & \multirow{2}{*}{HP. Prop.} & \multicolumn{3}{c|}{Compression Ratio} & \multicolumn{4}{c}{LMQA} \\
    \cmidrule{4-10}
        
        &  &  & HP. & LP. & Overall & HotpotQA & 2WikiMQA & Musique & Average \\
    \midrule
    Token w/o SC & Uniform & N.A. & N.A. & N.A. & 8 $\times$ & 33.81 & 38.76 & 12.56 & 28.38 \\ 
    \midrule
    \multirow{2}{*}{Token w. SC} & \multirow{2}{*}{Likelihood} & \multirow{4}{*}{1: 16} & 1 $\times$ & $\sim$ 16 $\times$ & 8 $\times$ & 33.18 & 37.80 & 11.82 & 27.60\\ 
    & & & 2 $\times$ & $\sim$ 11 $\times$ & 8 $\times$ & 35.17 & 38.60 & 11.84 & 28.54\\
    
    
    \cmidrule{1-2} 
    \cmidrule{4-10}
    \multirow{2}{*}{Sentence w. SC} & \multirow{2}{*}{Embedding} &  & 1 $\times$ & $\sim$ 16 $\times$ & 8 $\times$ & \textbf{36.30} & 39.15 & \textbf{14.33} & \textbf{29.93}\\
    & & & 2 $\times$ & $\sim$ 11 $\times$ & 8 $\times$ & 35.70 & \textbf{39.96} & 12.52 & 29.39\\
    \bottomrule
    \end{tabular}
    }
    \caption{Flexibility Analysis on LMQA using different importance estimators and compression ratios (HP/LP: high-priority / low-priority contexts. In our experiment, the top 1/16 (1:16) of the retrieved contexts are allocated with the high priority).}  
    \vspace{-15pt}
    \label{tab:estimator}
\end{table*}

\subsubsection{Efficiency}
We further explore the efficiency of FlexRAG using Torch Profiler~\cite{profiler}. 
The evaluation takes place on one NVIDIA A800 GPU, with a batch size of 8 with BF16 precision. FlexRAG is compared with Llama (w. retrieval), where both methods employ the identical input and output lengths. The experiment is conducted with the ODQA datasets, where EM is the quality metric. The compression ratio is varied from 2$\times$ to 8$\times$ for FlexRAG.  

As shown in Table~\ref{tab:efficiency}, FlexRAG is faster than Llama (w.r.) while improving the RAG's performance simultaneously. With the increasing of compression ratio, FlexRAG becomes even faster, leading to 3.54$\times$ reduction in CUDA time and 3.09$\times$ reduction in TFLOPs at a 16$\times$ compression ratio. Moreover, FlexRAG substantially improves upon Llama (w.r.) in terms of RAG quality, achieving a 10\% improvement in ODQA datasets at a 4$\times$ compression ratio.



\subsection{Flexibility Analysis}

\subsubsection{Flexibility in compression ratio}
We make evaluation of four compression ratios: $1\times$ (a special case without down-sampling), $2\times$, $4\times$, $8\times$. Both FlexRAG w. SC and FlexRAG w/o SC are included in our experiment, allowing us to evaluate the impact where selective compression is enabled or disabled. The experiments are performed on both LMQA and ODQA datasets. For LMQA, FlexRAG's compressed contexts may still exceed Llama-2's window size in some cases (unless $8\times$); as a result, truncation will be used when it has to. We utilize Llama (w/o retrieval) and Llama (w. retrieval) as the baselines for comparison. The experiment results are shown in Table \ref{tab:comp_ratio}, where the following observations can be made. 

First, FlexRAG consistently outperforms the baselines across all compression ratios. Even without selective compression, FlexRAG w/o SC already surpasses all baselines. When selective compression is enabled, FlexRAG w. SC further extends its empirical advantage. Such an observation indicates that the diverse compression ratios are effectively supported by FlexRAG across different use cases. Besides, it's worth noting that when the compression ratio is $1\times$, FlexRAG w/o SC receives the same truncated contexts as Llama (with retrieval). Therefore, any improvement it shows over Llama (with retrieval) provides a direct measure of FlexRAG's enhancement on RAG performance. 

Second, FlexRAG's performance is consistently improved on LMQA datasets when the compression ratio grows. The optimal result is achieved when the highest compression ratio $8\times$ is used. In contrast, FlexRAG exhibits an opposite tendency on ODQA datasets, where the lowest compression ratio $1\times$ presents the optimal result. As introduced, the retrieved contexts on LMQA exceed the window size of Llama-2; therefore, higher compression ratios help to bring in more information. However, the retrieved contexts on ODQA is concise; as a result, it will not introduce extra information with higher compression ratios, but only increase the information loss for the well-presented contexts. 




\begin{table*}[h]
    \centering
    \resizebox{0.80\linewidth}{!}{
    \begin{tabular}{l|ccccc|c}
    \toprule
        Retriever & BM25 & LLM-Embedder & BGE-base & BGE-large & E5-large & Average \\
    \midrule
        Llama (w. retrieval) & 31.20 & 39.22 & 35.88 & 37.25 & 39.82 & 36.67 \\
        FlexRAG & 44.58 & 46.75 & 45.80 & 45.95 & 48.47 & 46.31\\
    \bottomrule
    \end{tabular}
    }
    \caption{Average performance on ODQA datasets with
various retrievers.} 
    \label{tab:retriever}
\end{table*}

\subsubsection{Flexibility in compression method}
We make exploration of three alternative compression methods in our experiment. 1) Token w/o SC: this method is identical to FlexRAG w/o SC, operating at the token level with uniform down-sampling at an interval of 8 tokens. 2) Token w. SC: this method performs selective compression through token-level down-sampling, utilizing the likelihood-based importance estimator described in Eq. \ref{eq:likelihood}. 3) Sentence w. SC: this method performs selective compression via sentence-level down-sampling, employing the embedding-based importance estimator as outlined in Eq. \ref{eq:embedding}. We make further variation for the compression ratio: with an uniform overall compression ratio $8\times$ and high-priority proportion 1:16, we use two alternative sets of compression ratios for the high-priority (HP.) and low-priority (LP.) group. One is (HP: $1\times$, LP: $16\times$), also the default setting in our experiment; the other one is (HP: $2\times$, LP: $11\times$). The following observations can be made from the experiment results in Table \ref{tab:estimator}.

First, FlexRAG consistently maintains competitive performance across various scenarios. Notably, Sentence w. SC delivers the best results in our experiments. In contrast, Token w. SC shows sub-optimal performance, only outperforming Token w/o SC in specific cases. As previously discussed, token-level down-sampling can lead to incoherence, which diminishes the effectiveness of selective compression, especially under higher compression ratios. 

Second, FlexRAG exhibits varied performance given different allocation of compression ratios between the high-priority (HP) and low-priority (LP) groups. HotpotQA and Musique prefer the default allocation (HP: $1\times$, LP: $16\times$), while 2WikiMQA benefits more from a different setup (HP: $2\times$, LP: $11\times$). With a constrained overall compression ratio, assigning lower compression ratios to the HP group helps to preserve crucial information. However, this approach can also lead to the omission of less salient but still important content. Thus, finding the optimal allocation of compression ratios is about striking a delicate balance between these competing factors. 

\begin{figure}[t]
    \centering
    \includegraphics[width=1.00\columnwidth]{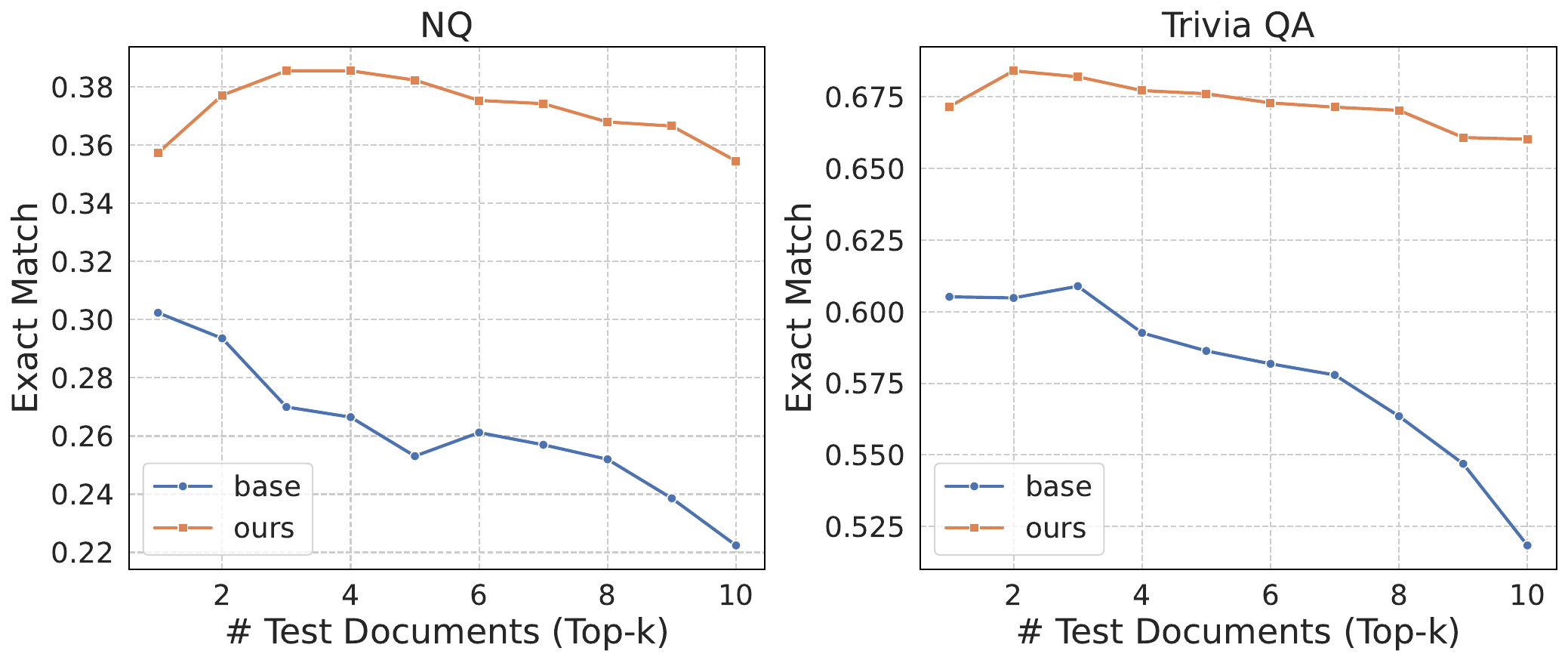}
    \caption{Performance on NQ and TriviaQA with varied \#Doc.}
    \label{fig:doc_num}
\end{figure}




\vspace{-5pt}
\subsection{Extended Analysis}

\subsubsection{Robustness to working conditions}
We first analyze the impact of using different retrievers. Beyond BGE-large, which was applied for both training and testing, we explore several alternative retrievers for testing, including BM25 \cite{robertson2009probabilistic}, LLM-Embedder \cite{zhang2023retrieve}, BGE-base \cite{bge_embedding}, and E5-large \cite{wang2022text}. The results are presented in Table~\ref{tab:retriever}, leading to the following key observations. First, FlexRAG consistently outperforms llama (w. retrieval) across all retrievers tested. Notably, this includes not only BGE-large, but also other retrievers differ from the one employed during training, demonstrating the strong generalizability of FlexRAG. Second, FlexRAG already achieves a superior performance even with a relatively weaker retriever, e.g., BM25. Additionally, its performance improves further when paired with stronger retrievers, like LLM-Embedder and E5-large. 

We further investigate the effect of the number of retrieved documents. In our experiment, we vary the number of retrieved contexts from the top 1 to the top 10 documents returned by the retriever. The results, shown in Figure~\ref{fig:doc_num}, lead to the following observations. First, our method (FlexRAG) consistently outperforms the baseline (Llama w. retrieval), and it demonstrates a greater stability, indicating that FlexRAG effectively handles variations in the number of retrieved documents. Second, FlexRAG's performance improves as the number of retrieved documents increases from 1 to 5, knowing that more useful information can be continually introduced. However, beyond this threshold, no further benefits are observed. According to previous studies~\cite{yoran2023making}, this plateau can be attributed to increased noise from irrelevant documents. Notably, FlexRAG experiences a much smaller performance decline compared to the baseline, suggesting it is more robust to noise. 

\begin{table}[t]
    \centering
    \resizebox{\columnwidth}{!}{
    \begin{tabular}{cc|cc}
    \toprule
    Factor & Setting & LMQA & ODQA \\
    \midrule
    \multirow{3}{*}{Training stage} & w/o pre-training & 22.32 & 39.91\\
    & w/o rag fine-tuning & 24.90 & 39.18\\
    & default setting* & \textbf{29.93} & \textbf{45.37}\\
    \midrule
    \multirow{3}{*}{Encoder Arch.} & first 4 layer & 25.07 & 39.85 \\
    & first 12 layer & 26.28 & 40.46\\
    & first 8 layer* & \textbf{29.93} & \textbf{45.37} \\
    \bottomrule
    \end{tabular}
    }
    \caption{Ablation studies of FlexRAG on LMQA and ODQA datasets. Default settings are marked with ``*''.}
    \label{tab:ablation}
\end{table}




\subsubsection{Ablation studies}
We first examine the significance of the two-stage training paradigm. In addition to the default method where both training stages are applied, we assess the impact of using only one of the stages: either w/o pre-training (i.e., RAG fine-tuning only) or w/o fine-tuning (i.e., pre-training only). The results, presented in Table \ref{tab:ablation}, indicate that pre-training with unlabeled data (w/o fine-tuning) significantly boosts FlexRAG's performance, as this stage alone already delivers competitive results. The subsequent RAG fine-tuning further enhances the performance, with the default two-stage method achieving the best results in the experiment. 

Next, we evaluate the architecture of the compressive encoder, with three alternatives tested: the first 4 layers of Llama-2, the first 8 layers of Llama-2 (default), and the first 12 layers of Llama-2. As shown in Table \ref{tab:ablation}, the best performance is achieved when the encoder uses the first 8 layers of Llama-2 as its backbone. In comparison, the smaller encoder (first 4 layers) is constrained by its limited expressiveness, while the larger encoder (first 12 layers) introduces greater disparity with the input layer of the downstream LLM. Therefore, selecting the appropriate encoder architecture requires balancing these considerations.

\section{Conclusion And Future Work}
In this paper, we have presented FlexRAG which brings forth compressed and optimized contexts for RAG tasks. FlexRAG's architecture allows for flexible production of compressed contexts across various compression ratios, while also supporting selective compression to preserve critical information. By leveraging a two-stage training workflow, FlexRAG effectively utilizes diverse training data, resulting in significant performance optimization. Our experiments across multiple QA datasets demonstrate FlexRAG’s cost-effectiveness, flexibility, and generalizability in different working conditions. Building on the progress of this preliminary work, future research will explore broader applications with more extensive LLM backbones and RAG tasks beyond question answering.



\bibliographystyle{ACM-Reference-Format}
\bibliography{reference}

\end{document}